\documentclass[10pt,twocolumn,letterpaper]{article}
\usepackage[table,xcdraw]{xcolor}
\usepackage{cvpr}
\usepackage{times}
\usepackage{epsfig}
\usepackage{graphicx}
\usepackage{amsmath}
\usepackage{amssymb}
\usepackage{multirow}

\usepackage[breaklinks=true,bookmarks=false,colorlinks=true]{hyperref}

\cvprfinalcopy % *** Uncomment this line for the final submission

\def\cvprPaperID{****} % *** Enter the CVPR Paper ID here

\ifcvprfinal\fi
\begin{document}

%%%%%%%%% TITLE
\title{Person re-identification with fusion of hand-crafted and deep pose-based body region features}

%% omit the following lines up until the closing ``}''.
%% Additional authors and addresses can be added with ``\and'',
%% just like the second author.
%% To save space, use either the email address or home page, not both
%\and
\author{Jubin Johnson\textsuperscript{1} \and Shunsuke Yasugi\textsuperscript{2} \and Yoichi Sugino\textsuperscript{2} \and Sugiri Pranata\textsuperscript{1}  \and Shengmei Shen\textsuperscript{1}\\
	\hspace*{-2.5cm}\textsuperscript{1}Panasonic R\&D Center   \hspace*{4cm}        \textsuperscript{2}Panasonic Corporation \\ \hspace*{1.2cm}Singapore \hspace*{3.5cm}Core Element Technology Development Center\\
	     \hspace*{7cm}       Japan\\
	{\tt\small \href{http://www.prdcsg.panasonic.com.sg/}{http://www.prdcsg.panasonic.com.sg/}}
	}

\maketitle
%\thispagestyle{empty}

%%%%%%%%% ABSTRACT
\begin{abstract}
   Person re-identification (re-ID) aims to accurately retrieve a person from a large-scale database of images captured across multiple cameras. Existing works learn deep representations using a large training subset of unique persons. However, identifying unseen persons is critical for a good re-ID algorithm. Moreover, the misalignment between person crops to detection errors or pose variations leads to poor feature matching. In this work, we present a fusion of handcrafted features and deep feature representation learned using multiple body parts to complement the global body features that achieves high performance on unseen test images. Pose information is used to detect body regions that are passed through Convolutional Neural Networks (CNN) to guide feature learning. Finally, a metric learning step enables robust distance matching on a discriminative subspace. Experimental results on 4 popular re-ID benchmark datasets namely VIPer, DukeMTMC-reID, Market-1501 and CUHK03 show that the proposed method achieves state-of-the-art performance in image-based person re-identification.        
\end{abstract}

%%%%%%%%% BODY TEXT
\section{Introduction}
Person re-ID is an important task in video surveillance and has various applications in security and law-enforcement. The main goal in re-ID is to retrieve all instances of a \emph{probe} person from a large \emph{gallery} set. In the instance of a security breach or event, law enforcement can use a re-ID system to automatically identify and track persons-of-interest, which would save hundreds of man-hours.     

Apart from the obvious challenges posed by viewpoint variations and occlusions, re-ID can be regarded as a zero-shot learning problem where the training and testing classes are different. It is therefore, crucial to learn discriminative representations for unseen persons. Existing approaches handle this challenge by either learning discriminative subspaces/metrics~\cite{yu2017cross,liao2015person,zheng2013reidentification,lisanti2015person,bai2017scalable,pedagadi2013local,chen2015similarity,zhang2016learning,peng2016unsupervised}, or generating discriminative features~\cite{matsukawa2016hierarchical,yang2014salient,farenzena2010person,zhao2013unsupervised,zheng2015query}. The success of deep learning in computer vision has led to emergence of works that learn discriminative features using CNNs~\cite{li2014deepreid,varior2016siamese,wu2016enhanced,zheng2017unlabeled,zheng2017person,zheng2017discriminatively} and through metric learning losses such as contrastive loss~\cite{varior2016gated}, triplet loss~\cite{liu2017end,hermans2017defense}, and more recently quadruplet loss~\cite{chen2017beyond}. 

\begin{figure}[t]
	\centering
	\includegraphics[clip, trim=7.8cm 6.0cm 13.8cm 3.9cm, width=0.5\textwidth]{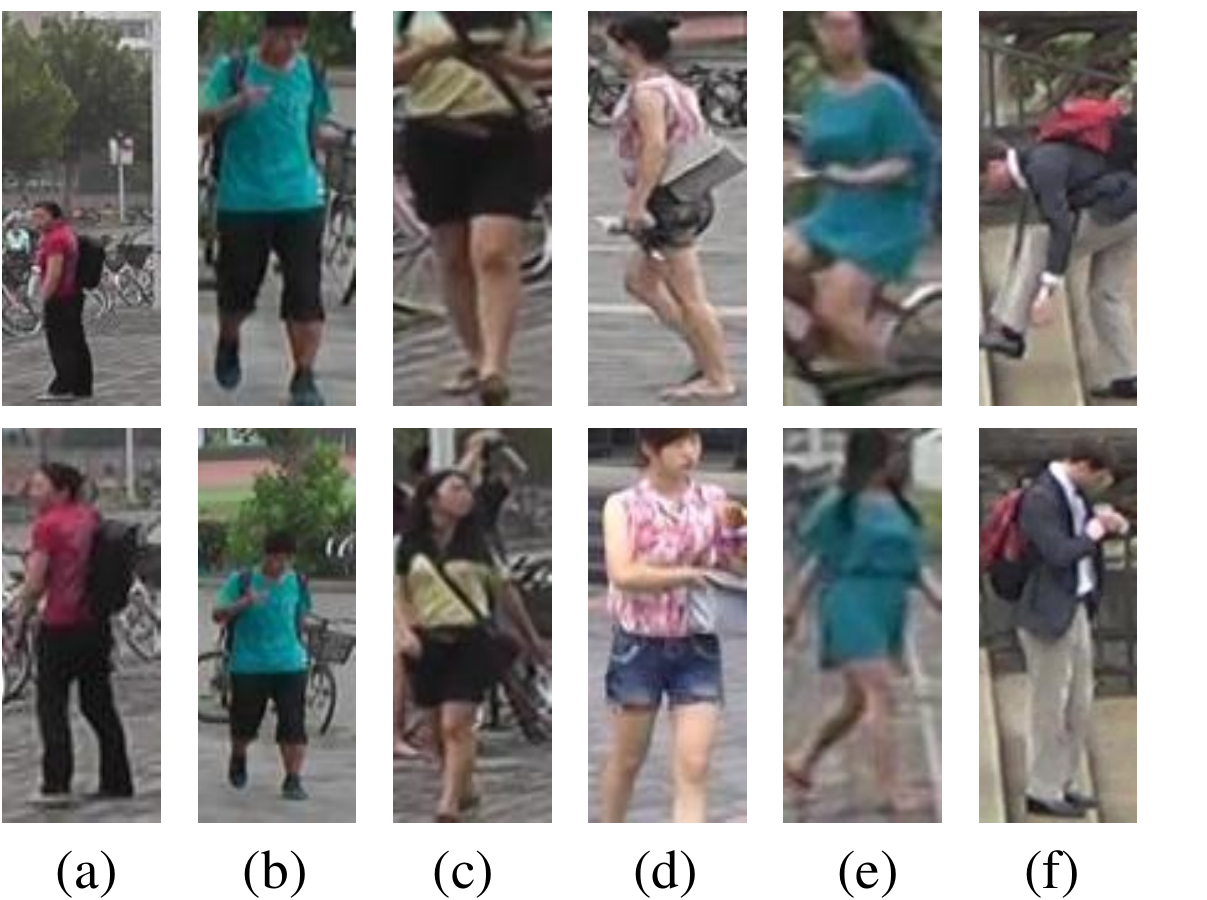}
	\caption{Common challenges in image-based re-ID. (a-b) Scale variations, (c-d) detection errors, (e-f) pose misalignments}
	\label{fig:introfig}
\end{figure}
A majority of these approaches learn a global feature representation for the person crop, without considering the spatial information contained in the image. Previous works~\cite{yao2017deep} have shown that the representations learned by a deep classification model on the global image are focused mainly on one body region- the upper body. The drawbacks of such global representations are 
 illustrated in Fig.~\ref{fig:introfig}. Since person crops for large-scale datasets like Market1501 and CUHK03 are generated from video frames using detectors such as DPM~\cite{felzenszwalb2010object}, inaccurate detection boxes might impact feature learning, e.g., Fig.~\ref{fig:introfig} (a-d).  The top half of the images in Fig.~\ref{fig:introfig}(a) correspond to the head-shoulder region and background respectively. Directly comparing the feature maps learned from the global image would impede the similarity score. The pose change or non-rigid body deformation makes the metric learning difficult, e.g., Fig.~\ref{fig:introfig} (e-f). Moreover, occluded parts of the human body might introduce irrelevant context into the learned feature and it is non-trivial to emphasize local differences in a global feature, especially
when we have to distinguish two people with very similar appearances. To explicitly overcome these drawbacks, recent studies have paid attention to part-based, local feature learning. Some works \cite{varior2016siamese,yao2017deep} divide the whole body into a few fixed parts, without considering the alignment between parts. However, it still suffers from inaccurate detection box, pose variation, and occlusion. 

In the proposed re-ID framework, pose information is utilized to guide the feature learning process for various body parts. Pose has been recently used to guide local learning for better alignment of features~\cite{zhao2017spindle,su2017pose,zheng2017pose}. Our work is inspired by \cite{zhao2017spindle} where multiple body regions are used to guide the feature learning followed by a tree-based feature fusion step to effectively combine the different local features. We show that such detailed level of granularity in the body parts is not always beneficial to the learning process. A simple concatenation of the features learned from three body regions - head , body and leg, along with the global features generalize well across datasets captured from different domains. Second, we adopt a disjoint framework which not only
utilizes the powerful feature extraction capability of CNN but also brings into
play the exclusive superiority of complementary handcrafted features and advanced metric learning method.       

\section{Related Work}
We review work closely related to our method. An extensive survey of person re-ID can be found at \cite{zheng2016person,karanam2016comprehensive}. Most ReID pipelines are composed of two main steps,
feature learning and metric learning. For feature learning, several effective approaches have been proposed,
for example, the Ensemble of Local Features (ELF)~\cite{gray2008viewpoint}, the Local Maximal Oc-
currence (LOMO)~\cite{liao2015person}. These handcrafted features have made impressive improve-
ments over robust feature representation, and advanced the person re-id research. The promising performance of CNN on ImageNet classification
indicates that classification network extracts discriminative
image features. Therefore, several works \cite{xiao2016learning,zheng2017person,wu2016enhanced} fine-tuned the classification networks on target
datasets as feature extractors for person re-ID. For example,
Xiao~\etal~\cite{xiao2016learning} propose a novel dropout strategy to train a
classification model with multiple datasets jointly. Wu \etal~\cite{wu2016enhanced} combine the hand-crafted histogram features and
deep features to fine-tune the classification network.
Besides classification network, siamese network and triplet network are two other popular networks for person ReID.
The siamese network takes a pair of images as input, and
is trained to verify the similarity between those two images \cite{ahmed2015improved,zheng2017discriminatively,varior2016siamese,shi2016embedding}. Ahmed \etal~\cite{ahmed2015improved} and Zheng \etal~\cite{zheng2017discriminatively} employ the siamese network to infer the description
and a corresponding similarity metric simultaneously.
Shi \etal~\cite{shi2016embedding} replace the Euclidean distance with Mahalanobis
distance in the siamese network. Varior \etal~\cite{varior2016siamese}
combine the LSTM and siamese network for person ReID.
Some other works \cite{liu2017end,hermans2017defense} employ triplet networks to
learn the representation for person ReID. Recently, many works generate deep representation from
local parts \cite{su2017pose,zhao2017spindle,li2017learning,zhang2017alignedreid}. Li~\etal~\cite{li2017learning} employ
Spatial Transform Network (STN) for part localization,
and propose Multli-Scale Context-Aware Network to infer
representations on the generated local parts.  
\begin{figure}[h]
	\centering
	\includegraphics[clip, trim=0.5cm 2.2cm 0cm 1.0cm, width=0.5\textwidth]{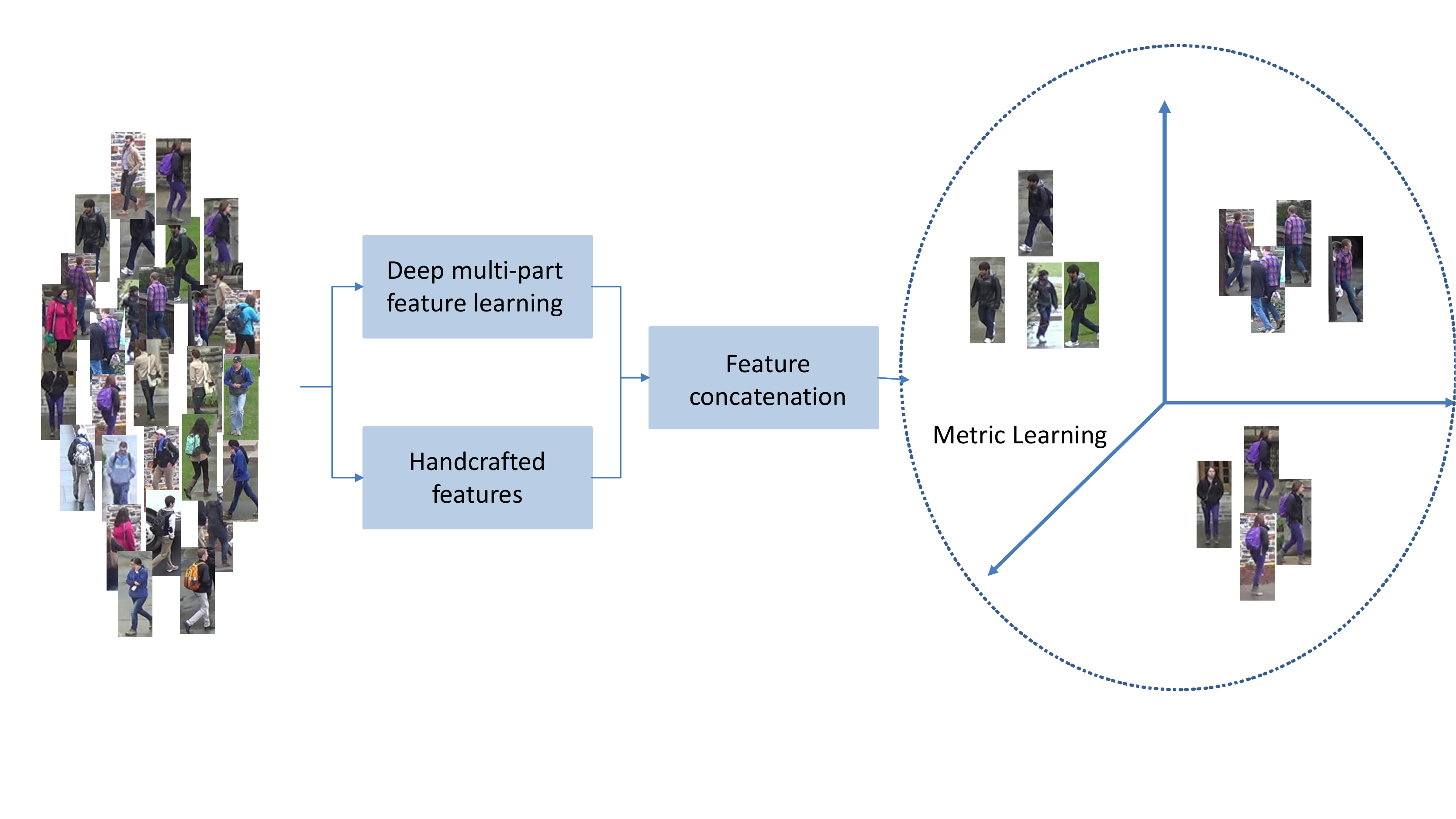}
	\caption{Framework of the proposed method}
	\label{fig:ovrframework}
\end{figure}

Zhang~\etal~\cite{zhang2017alignedreid} align the local features with the global features in a mutual learning framework. Pose invariant
embedding (PIE)~\cite{zheng2017pose} aligns pedestrians to a standard pose to
reduce the impact of pose variation. A Global-Local Alignment
Descriptor (GLAD)~\cite{wei2017glad} does not directly align
pedestrians, but rather detects key pose points and extracts
local features from corresponding regions.
\cite{zhang2017deep} presents a deep mutual learning
strategy where an ensemble of students learn collaboratively
and teach each other throughout the training process.
DarkRank~\cite{chen2017darkrank} introduces a new type of knowledge-cross
sample similarity for model compression and acceleration,
achieving state-of-the-art performance. These methods use
mutual learning in classification

Our work is inspired by the part-based approach introduced by Zhao \etal~\cite{zhao2017spindle}. They firstly extract human body parts with fourteen
body joints, then fuse the features extracted on body
parts using a feature fusion network that is trained end-to-end. We argue that the feature fusion network and the micro-body regions actually hamper the performance in test images. Experimental evaluations are provided in section \ref{sec:experimetal} to validate this claim.
\begin{figure*}[ht!]
	\centering
	\includegraphics[clip, trim=0.5cm 0.2cm 1.5cm 1.2cm, width=1\textwidth]{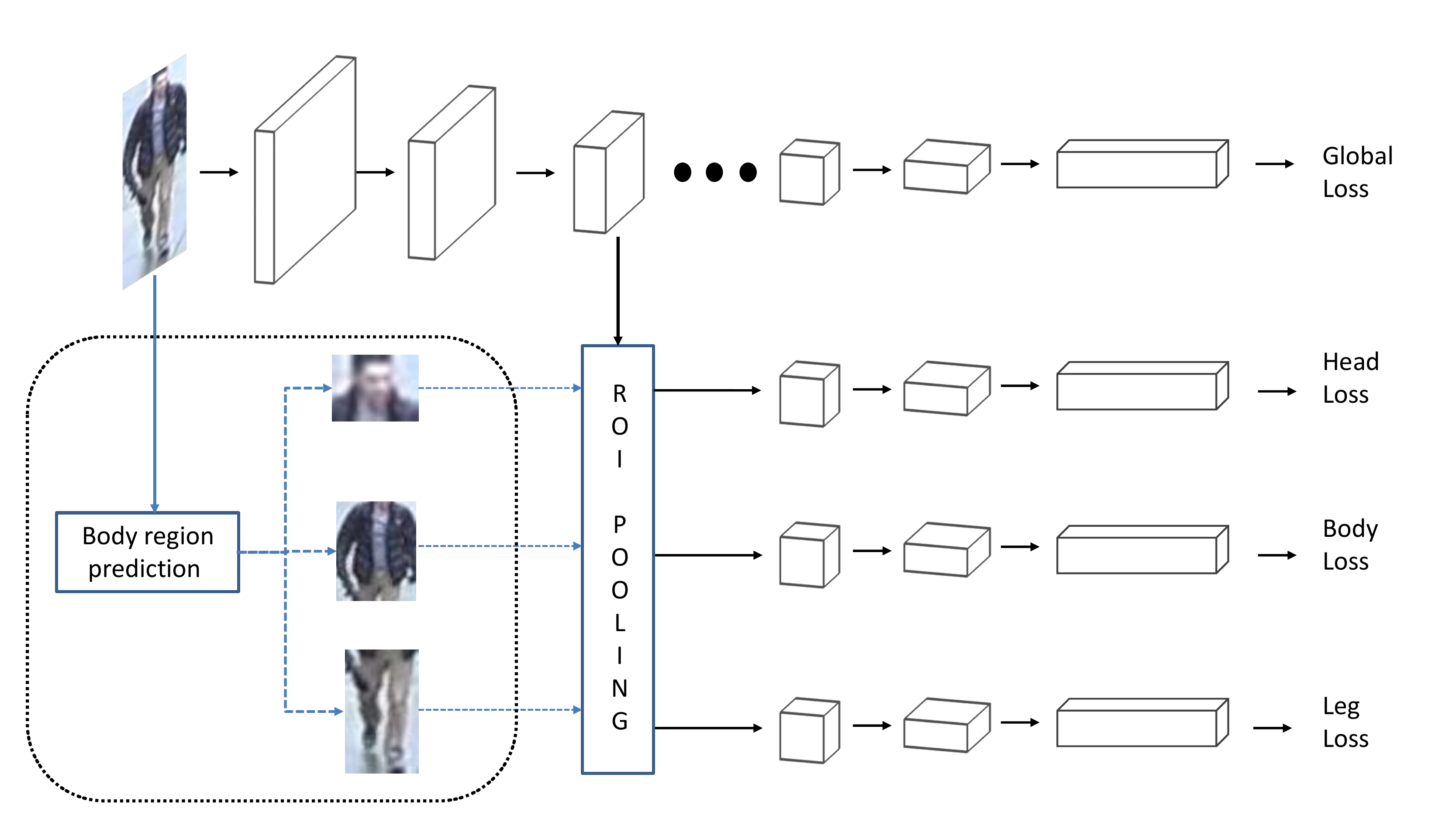}
	\caption{Illustration of the body part-based deep feature learning}
	\label{fig:framework}
\end{figure*}
\section{Proposed Method}
The proposed re-ID framework involves two stages for learning the feature representation of an input person image. The overall framework is illustrated in Fig.~\ref{fig:ovrframework}. First, the three body regions, namely head, upper body and lower body are learned using a pose prediction network. The body regions are used as guidance to the network for learning region-specific deep features in the next step through a ROI pooling layer as illustrated in Fig.~\ref{fig:framework}. The handcrafted features are learned and combined with the deep features in the next step. Finally, metric learning is used to generate a discriminative subspace for accurate comparison between the query and probe features.

\subsection{Body region prediction}
\begin{figure}[h]
	\centering
	\includegraphics[clip, trim=1.4cm 5.4cm 22.6cm 3.0cm, width=0.45\textwidth]{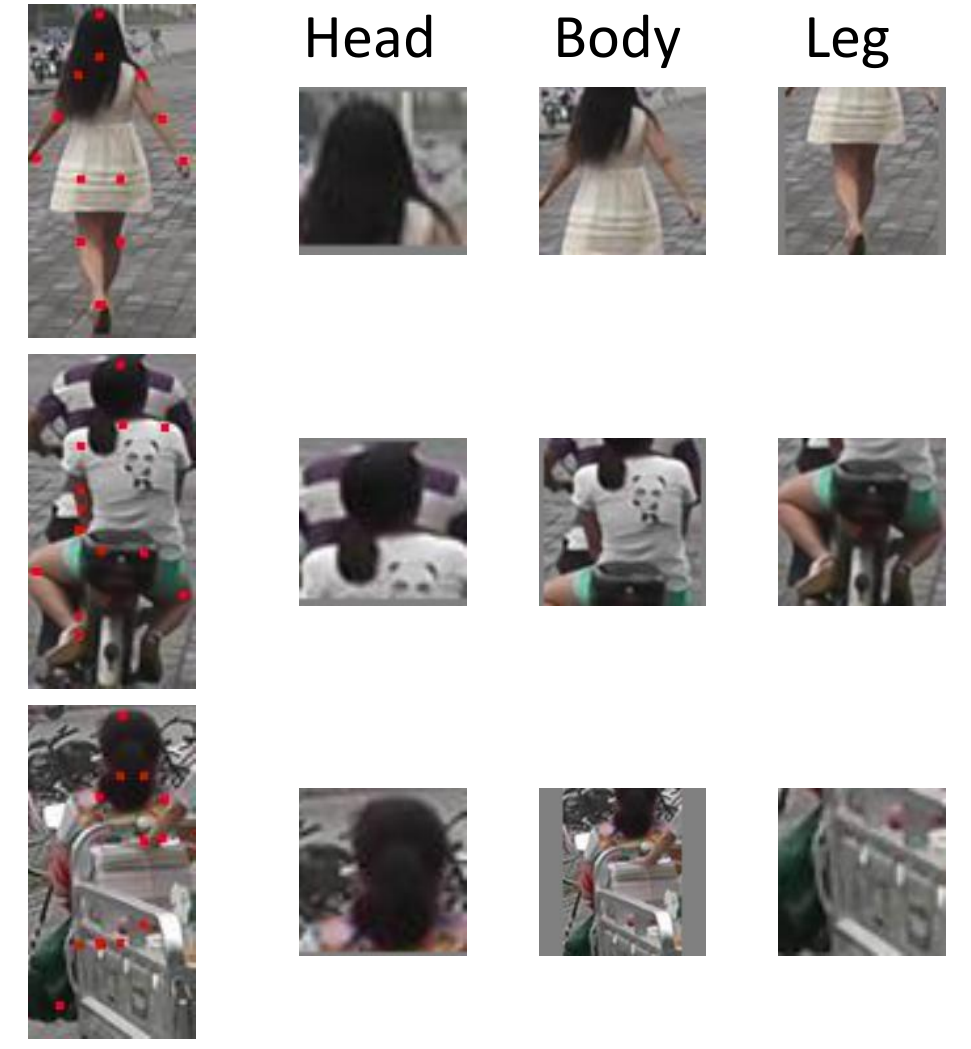}
	\caption{Examples of person images and the corresponding body regions extracted from them.}
	\label{fig:bodyregions}
\end{figure}

Similar to Zhao \etal~\cite{zhao2017spindle}, human body region information is extracted using a pose prediction network (PPN) which will be used to guide the feature learning process for each sub-region. A sequential framework inspired by the Convolutional Pose Machines (CPM)~\cite{wei2016convolutional} generates body joint 
response maps in a coarse-to-fine manner. In each stage, a
convolutional network is used to extract image features and
then combine the response maps from the previous stage,
to produce increasingly refined estimations for body joint
locations.  Given an 
input image, the PPN predicts fourteen response maps $F=\mathbb{R}^{W^{'}\times H^{'}\times 14}$. The body joint locations are then obtained by max pooling each of the response maps, followed by grouping them to generate three rectangle region proposals
representing the head, the upper-body and lower-body regions, respectively. A few examples are shown in Fig.~\ref{fig:bodyregions}. More details can be found in \cite{zhao2017spindle} with the exception that we do not mine out micro body regions.

\subsection{Multi-part representation learning for re-ID}
The body parts generated in the previous step guides the learning for the multi-part representation The general flowchart for training the network is shown in
Fig.~\ref{fig:framework}. The network takes the person image together with the
3 body regions as input and computes one global feature
and three part features. Details of the network and structure are present in \cite{zhao2017spindle}. It essentially contains three inception modules and an ROI pooling stage for pooling out the body region features.   Each feature is 256 dimensional which is concatenated to obtain the  final feature vector. During testing this multi-part feature is extracted and used
to distinguish different persons.
\subsection{Fusion of handcrafted features and metric learning} 
  In addition to the deep features, low-level features like color texture and illumination are crucial factors for describing images of persons. In addition to HSV color histogram, the Scale Invariant Local Ternary
  Pattern (SILTP) descriptor is utilized for illumination invariant texture de-
  scription. In order to extract spatial information, sliding windows (subwindow,
  i.e., 10×10 size) are applied to describe local details of a person image. Adopting
  pyramid representation, within each subwindow, we extract two scales of SILTP
  histograms and a joint HSV histogram. To address viewpoint changes, we check
  all subwindows at the same horizontal location, and maximize the local occur-
  rence of each pattern (i.e., the same histogram bin) among those subwindows,
  it is called Local Maximal Occurrence (LOMO) features. For 128$\times$48 image,
  all the computed local maximal occurrences are concatenated into the final de-
  scriptor with 26960 dimensions and it is called LOMO. Finally, a log transform
  is applied to suppress large bin values, and both HSV and SILTP features are
  normalized. For metric learning, we adopt the widely used Cross-view Quadratic Discriminant Analysis (XQDA)~\cite{liao2015person} to learn the subspace.

\section{Experiments}
\label{sec:experimetal}
\subsection{Datasets}
The proposed framework is evaluated on the most widely used datasets in the past year for image-based re-ID namely, \emph{VIPeR}~\cite{gray2007evaluating}, \emph{CUHK-03}~\cite{li2014deepreid}, \emph{Market-1501}~\cite{zheng2015scalable} and \emph{DukeMTMC-reID}~\cite{ristani2016performance,zheng2017unlabeled}.

\textbf{VIPeR} is one of the earliest and widely used datasets for person re-ID. It contains 632 unique identities captured across two cameras, with each identity having one image per camera. The dataset is split randomly into equal halves for training and testing.

\textbf{CUHK-03} consists of 14,097 cropped images from
1,467 identities. For each identity, images are captured from
two cameras and there are about 5 images for each view.
Two ways are used to produce the cropped images, i.e., human
annotation and detection by DPM~\cite{felzenszwalb2010object}. Our evaluation is based on the combined dataset. We use the standard experimental setting
to select 1,367 identities for training, and the rest 100 for
testing.

\textbf{Market-1501} contains 32,668 images from 1,501
identities, and each image is annotated with a bounding box
detected by DPM. Each identity is captured by at most six
cameras. We use the standard training, testing, and query
split provided by the authors in \cite{zheng2015scalable}. 

\textbf{DukeMTMC-reID} is a subset of the DukeMTMC~\cite{ristani2016performance} for image-based re-identification, in the format of the Market-1501 dataset. It contains 16,522 training images of 702 identities, 2,228 query images of the other 702 identities and 17,661 gallery images. We report mAP and Rank-1 precision on Market-1501 and DukeMTMC-reID datasets, while rank-1, -5, and -10 accuracies are reported on the other 2 datasets. Similar to \cite{xiao2016learning,zhao2017spindle,yao2017deep}, we learn a single model using the training splits of all the four datasets. The statistics of the training and testing splits are shown in Table~\ref{tab:splits}.

\begin{table}[t]
	\centering
	\caption{Details of the datasets and the corresponding training/testing splits.}
	\label{tab:splits}
	\resizebox{0.5\textwidth}{!}{%
		\begin{tabular}{lllll}
			\hline
			Dataset       & \#ID  & \#Train/Val img & \#Probe/Gallery ID & \#Probe/Gallery img \\ \hline \hline
			VIPeR         & 632  & 506/126        & 316/316           & 316/316            \\
			CUHK-03       & 1467 & 21012/5252     & 100/100           & 100/100            \\
			Market-1501   & 1501 & 10348/2588     & 750/750+junk      & 3368/19732         \\
			DukeMTMC-reID & 1404 & 13218/3304     & 702/702           & 2228/17661         \\ \hline \hline
		\end{tabular}%
	}
\end{table}
\begin{table}[h]
	\centering
	\caption{Comparisons on \textit{Market1501} in single query setting}
	\label{tab:marketresults}
	\resizebox{0.4\textwidth}{!}{%
		\begin{tabular}{l|ll}
			\hline
			Method                  & Rank-1 & mAP   \\ \hline \hline
			LOMO+XQDA~\cite{liao2015person}               & 43.79  & 22.22 \\
			k-reciprocal~\cite{zhong2017re}            & 77.11  & 63.63 \\
			PIE~\cite{zheng2017pose}                     & 79.33  & 55.95 \\
			Deep Context-aware~\cite{li2017learning}      & 80.31  & 57.53 \\
			Deep Part-Aligned~\cite{zhao2017deeply}       & 81.0   & 63.4  \\
			Scalable SSM~\cite{bai2017scalable}          & 82.21  & 68.80 \\
			SVDNet~\cite{sun2017svdnet}                  & 82.3   & 62.1  \\
			Pose-driven~\cite{su2017pose}             & 84.14  & 63.41 \\
			APR~\cite{lin2017improving}                     & 84.29  & 64.67 \\
			In Defense Triplet~\cite{hermans2017defense}      & 84.92  & 69.14 \\
			Deep Mutual Learning~\cite{zhang2017deep}    & 87.73  & 68.83 \\
			REDA~\cite{zhong2017random}                    & 87.08  & 71.31 \\
			SpindleNet~\cite{zhao2017spindle}              & 84.5   & 65.1  \\
			Darkrank~\cite{chen2017darkrank}                & 89.8   & 74.3  \\
			GLAD~\cite{wei2017glad}                    & 89.9   & 73.9  \\
			AlignedReID~\cite{zhang2017alignedreid}             & 91.8   & \textbf{79.3}  \\ \hline
			Ours                    & 91.9   & 78.39 \\
			Ours + LOMO +XQDA       & \textbf{92.09}  & 79.11 \\ \hline
		\end{tabular}%
	}
\end{table}

\begin{table}[h]
	\centering
	\caption{Comparisons on \textit{DukeMTMC-reID} in single query setting}
	\label{tab:dukemtmcresults}
	\resizebox{0.4\textwidth}{!}{%
		\begin{tabular}{l|ll}
			\hline
			Method                  & Rank-1 & mAP   \\ \hline \hline
			LOMO+XQDA~\cite{liao2015person}               & 30.75  & 17.04 \\
			IDE~\cite{zheng2016person}            & 65.22  & 44.99 \\
			GAN~\cite{zheng2017unlabeled}                     & 67.68  & 47.13 \\
			SpindleNet~\cite{zhao2017spindle}              & 68.9   & 46.2  \\
			Discriminatively~\cite{zheng2017discriminatively}      & 68.9  & 49.3 \\
			APR~\cite{lin2017improving}                     & 70.69 & 51.88 \\
			PAN~\cite{zheng2017pedestrian}             & 71.59  & 51.51 \\
			SVDNet~\cite{sun2017svdnet}                  & 76.7   & 56.8  \\

			DPFL~\cite{chen2017person}      & 79.2  & 60.6 \\
			PSE~\cite{sarfraz2017pose}                    & 79.8   & 62.0  \\
			REDA~\cite{zhong2017random}                    & 79.31  & 62.44 \\
			Mid-level~\cite{yu2017devil}    & 80.43  & 63.88 \\
			Deep-Person~\cite{bai2017deep}                & 80.90   & 64.80  \\
			
			PCB~\cite{sun2017beyond}             & 83.3   & 69.2  \\ 
			GP-reID~\cite{almazan2018re}             & \textbf{85.2}   & \textbf{72.8}  \\
			\hline
			Ours                    & 79.75   & 63.6 \\
			Ours + LOMO +XQDA       & 80.36  & 64.8 \\ \hline
		\end{tabular}%
	}
\end{table}
\subsection{Comparisons with other methods} 
We compare the performance of our proposed method on the standard evaluation protocols for the 4 datasets. The results are reported in Tables ~\ref{tab:marketresults} - \ref{tab:viperresults}. In the tables, \emph{Ours} represents our results using cosine similarity for matching, while \emph{Ours+LOMO+XQDA} represents the full framework with the addition of handcrafted features and metric learning.

On Market-1501, the proposed method achieves state-of-the-art performance with Rank-1 accuracy of 92.09\% as shown in Table~\ref{tab:marketresults}. Our mAP of 79.1\% is only slightly lower than 79.3\% of AlignedReID~\cite{zhang2017alignedreid}. The influence of  metric learning is not significant for datasets with large gallery sets, nevertheless, the inclusion does bring about a slight increase in performance. 

Table~\ref{tab:dukemtmcresults} shows the performance on DukeMTMC-reID, obtaining accuracy of 80.36\% and 64.8\% on Rank-1 and mAP measures. ,respectively. Although some recent methods~\cite{almazan2018re,bai2017deep,sun2017beyond} perform better than the proposed method on this dataset, they have lower scores on the other three datasets. This shows that our method is able to generalize well across datasets captured from different domains.

VIPeR and CUHK-03 follow the single-shot evaluation strategy, where only one query and gallery image is selected for each identity. Hence, we use the CMC ranking evaluation metric to compare with other methods. The comparisons on CUHK-03 are summarized in Table~\ref{tab:cuhk03results}. The proposed method obtains accuracy of 92.7\%, 99.1\% and 99.6\% on Rank- 1, 5 and 10 measures which is the state-of-the-art. We even outperform methods such as ~\cite{zhao2017spindle,liu2017hydraplus} which use multiple images of the same identity in the gallery. 

On VIPeR dataset, we achieve 63.6\% rank-1 accuracy, which is around 7\% more than the current state-of-the-art. As can be seen from Table~\ref{tab:viperresults}, for small-scale datasets, a big factor for this improvement is the metric learning step.      
\begin{table}[h]
	\centering
	\caption{Comparisons on \textit{CUHK03} dataset}
	\label{tab:cuhk03results}
	\resizebox{0.4\textwidth}{!}{%
		\begin{tabular}{llll}
			& \multicolumn{3}{c}{Rank} \\
			\hline
			\multicolumn{1}{l|}{Method}                  & r=1 & r=5 & r=10    \\ \hline \hline
			\multicolumn{1}{l|}{LOMO+XQDA~\cite{liao2015person}}               & 44.6  & - & - \\
			\multicolumn{1}{l|}{DNS~\cite{zhang2016learning}}            & 62.6  & 90.0 & 94.8 \\
			\multicolumn{1}{l|}{Gated SCNN~\cite{varior2016gated}}                     & 61.8 & 88.1 & 92.6 \\
			\multicolumn{1}{l|}{k-reciprocal~\cite{zhong2017re}}        & 64.0   & - & -  \\
			\multicolumn{1}{l|}{Scalable SSM~\cite{bai2017scalable}}          & 76.6  & 94.6 & 98.0 \\
			\multicolumn{1}{l|}{PL-Net~\cite{yao2017deep}}      & 82.8  & 96.6 & 98.6 \\
			\multicolumn{1}{l|}{Deep Part-Aligned~\cite{zhao2017deeply}}       & 85.4   & 97.6 & 99.4  \\
			\multicolumn{1}{l|}{GAN~\cite{zheng2017unlabeled}}                     & 84.6  & 97.6 & 98.9 \\
			\multicolumn{1}{l|}{Discriminatively~\cite{zheng2017discriminatively}}      & 83.4  & 97.1 & 98.7 \\
			\multicolumn{1}{l|}{In Defense Triplet~\cite{hermans2017defense}}      & 75.5  & 95.2 & 99.2 \\
			\multicolumn{1}{l|}{HydraPlus-Net~\cite{liu2017hydraplus}}    & 91.8  & 98.4 & 99.1 \\
			\multicolumn{1}{l|}{Pose-driven~\cite{su2017pose}}             & 88.7  & 98.6 & 99.2\\
			\multicolumn{1}{l|}{SpindleNet~\cite{zhao2017spindle}}              & 88.5   & 97.8  & 98.6 \\
			\multicolumn{1}{l|}{Darkrank~\cite{chen2017darkrank}}                & 89.7   & 98.4 & 99.2 \\
			\multicolumn{1}{l|}{GLAD~\cite{wei2017glad}}                    & 85.0   & 97.9 & 99.1  \\
			\multicolumn{1}{l|}{AlignedReID~\cite{zhang2017alignedreid}}             & 92.4   & 98.9 & 99.5  \\ \hline
			\multicolumn{1}{l|}{Ours}                    & 89.4   & 98.4 & 99.1 \\
			\multicolumn{1}{l|}{Ours + LOMO +XQDA}       & \textbf{92.7}  & \textbf{99.1} & \textbf{99.6} \\ \hline
		\end{tabular}%
	}
\end{table}

\begin{table}[h]
	\centering
	\caption{Comparisons on \textit{VIPeR} dataset}
	\label{tab:viperresults}
	\resizebox{0.4\textwidth}{!}{%
		\begin{tabular}{llll}
			& \multicolumn{3}{c}{Rank} \\
			\hline
			\multicolumn{1}{l|}{Method}                  & r=1 & r=5 & r=10    \\ \hline \hline
			\multicolumn{1}{l|}{Deepreid~\cite{li2014deepreid}}                     & 19.9  & 49.3 & 64.7 \\
			\multicolumn{1}{l|}{LOMO+XQDA~\cite{liao2015person}}               & 40.0  & - & 80.5 \\
			\multicolumn{1}{l|}{DNS~\cite{zhang2016learning}}            & 41.0  & 69.8 & 81.6 \\
			\multicolumn{1}{l|}{Gated SCNN~\cite{varior2016gated}}                     & 37.8 & 66.9 & 77.4 \\
			\multicolumn{1}{l|}{TMA~\cite{martinel2016temporal}}        & 48.2   & 87.7 & 93.5  \\
			\multicolumn{1}{l|}{Scalable SSM~\cite{bai2017scalable}}          & 53.7  & \textbf{91.5} & 96.1 \\
			
			\multicolumn{1}{l|}{Deep Part-Aligned~\cite{zhao2017deeply}}       & 48.7   & 74.7 & 85.1  \\
			
			\multicolumn{1}{l|}{MuSDeep~\cite{qian2017multi}}      & 43.0  & 74.4 & 85.8 \\
			
			\multicolumn{1}{l|}{Pose-driven~\cite{su2017pose}}             & 51.3  & 74.1 & 84.2\\
			\multicolumn{1}{l|}{SpindleNet~\cite{zhao2017spindle}}              & 53.8   & 74.1  & 83.2 \\
			\multicolumn{1}{l|}{GLAD~\cite{wei2017glad}}                    & 54.8   & 74.5 & 83.5  \\
			\multicolumn{1}{l|}{PL-Net~\cite{yao2017deep}}      & 56.7  & 82.6 & 90.0 \\
			\multicolumn{1}{l|}{SCSP~\cite{chen2016similarity}}             & 53.5   & \textbf{91.5} & \textbf{96.7}  \\
			\multicolumn{1}{l|}{HydraPlus-Net~\cite{liu2017hydraplus}}    & 56.6  & 78.8 & 87.0 \\ \hline
			\multicolumn{1}{l|}{Ours}                    & 45.6   & 66.1 & 74.7 \\
			\multicolumn{1}{l|}{Ours + LOMO +XQDA}       & \textbf{63.3}  & 83.5 & 91.1 \\ \hline
		\end{tabular}%
	}
\end{table}
\begin{table}[]
	\centering
	\caption{Investigation on the effect of each body part towards the final representation performance on Market-1501 dataset.}
	\label{tab:featurefusion}
	\resizebox{0.45\textwidth}{!}{%
		\begin{tabular}{lll}
			\hline
			Market-1501                              & Rank-1        & mAP           \\ \hline
			Global                                   & 86.4          & 67.6          \\
			Head                                     & 36.7          & 16.8          \\
			Body                                     & 62.9          & 38.3          \\
			Leg                                      & 60.4          & 36.2          \\
			Concat(Global + 7 body regions)          & 85.2          & 64.56         \\
			SpindleNet FFN (Global + 7 body regions) & 84.5          & 65.1          \\
			Concat(Global + 3 body regions)          & \textbf{88.3} & \textbf{69.5} \\ \hline
		\end{tabular}%
	}
\end{table}
\begin{table}[h]
	\centering
	\caption{Comparison of architecture in performance for DukeMTMC-reID and Market-1501}
	\label{tab:architecturecomp}
	\begin{tabular}{lll|ll}
		\hline
		\multirow{2}{*}{\begin{tabular}[c]{@{}l@{}}Backbone \\ architecture\end{tabular}} & \multicolumn{2}{c|}{DukeMTMC-reID} & \multicolumn{2}{c}{Market-1501} \\ \cline{2-5} 
		& Rank-1           & mAP             & Rank-1          & mAP           \\ \hline
		Inception                                                                         & 79.75            & 63.6            & 91.9            & 78.39         \\
		Resnet-50                                                                         & 77.69            & 61.13           & 91.56           & 75.63         \\ \hline
	\end{tabular}
\end{table} 
\subsection{Comparisons with feature fusion of SpindleNet} 
SpindleNet~\cite{zhao2017spindle} uses a tree fusion structure to combine the features obtained from the multiple body regions and showed that their fusion strategy based on feature competition and fine-tuning obtained the best performance. In our experiments, we observed that the addition of finer (micro) body regions deteriorated the performance of the overall framework. Table~\ref{tab:featurefusion} shows the experimental results validating this on the Market-1501 dataset. We trained SpindleNet and our network on the training split of this dataset. The addition of the micro body regions is shown to bring down the performance of both the feature fusion network as well as the linear concatenation. It is to be noted that the global features perform better than the FFN of SpindleNet. The use of the macro body regions - head leg and body, utilize complementary information that cannot be well represented by the global features, improving the performance by 2\% on both mAP and Rank-1. 

\begin{figure}[h!]
	{\footnotesize Query  \hspace*{1.8cm} Top 10 matches from gallery}\\
	{\centering
		\includegraphics[clip, trim=1.5cm 0.1cm 14.01cm 0.1cm, width=0.5\textwidth]{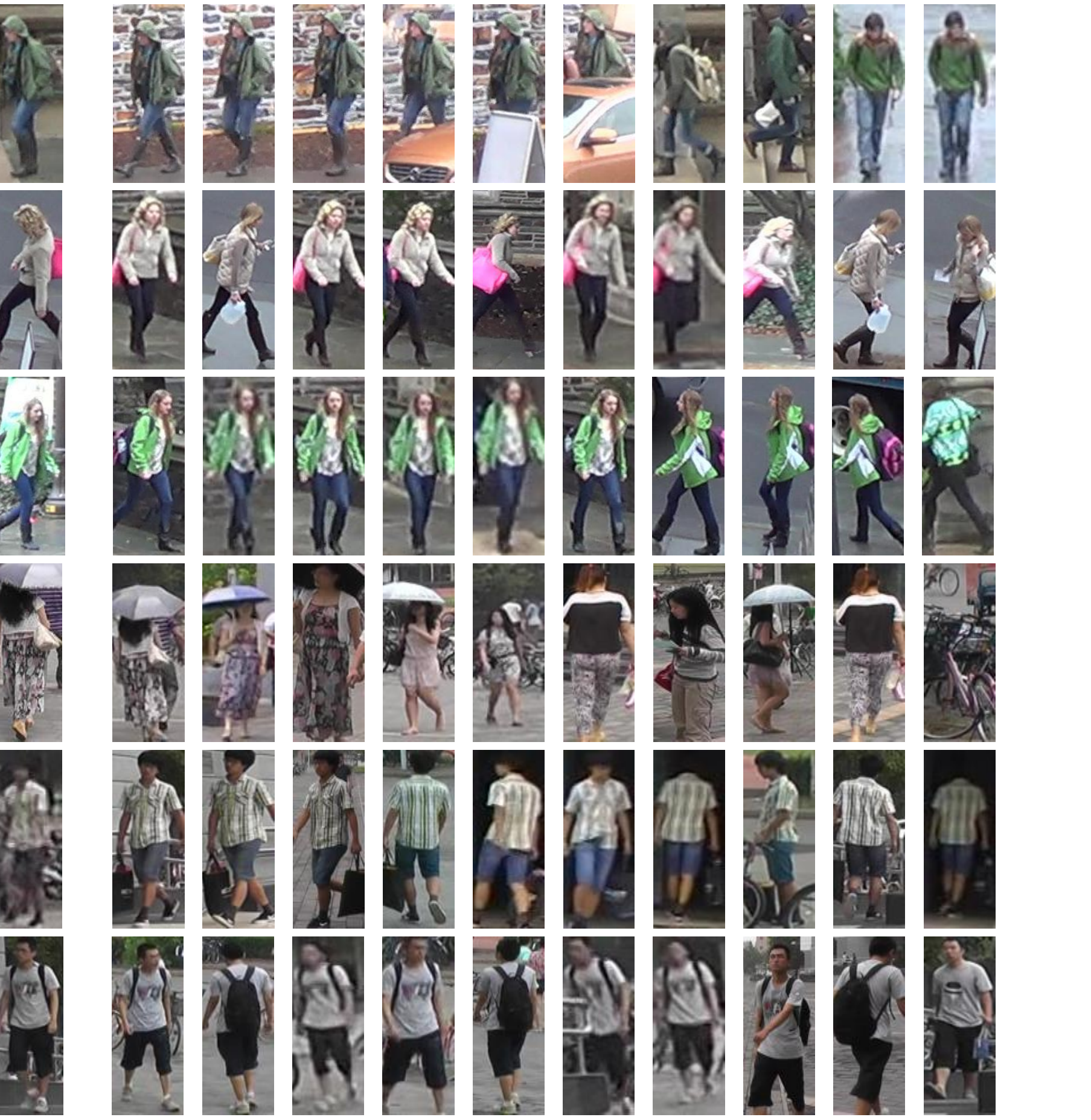}}
	\caption{Qualitative evaluation on DukeMTMC-reID (first three rows) and  Market1501 datasets. The first column is the probe image, followed by the top 10 results from the gallery.}
	\label{fig:marketdukequal}
\end{figure}
\begin{figure}[h!]
	{\footnotesize Query  \hspace*{1.8cm} Top 10 matches from gallery}\\
	{\centering
		\includegraphics[clip, trim=1.5cm 0cm 14.01cm 0.0cm, width=0.5\textwidth]{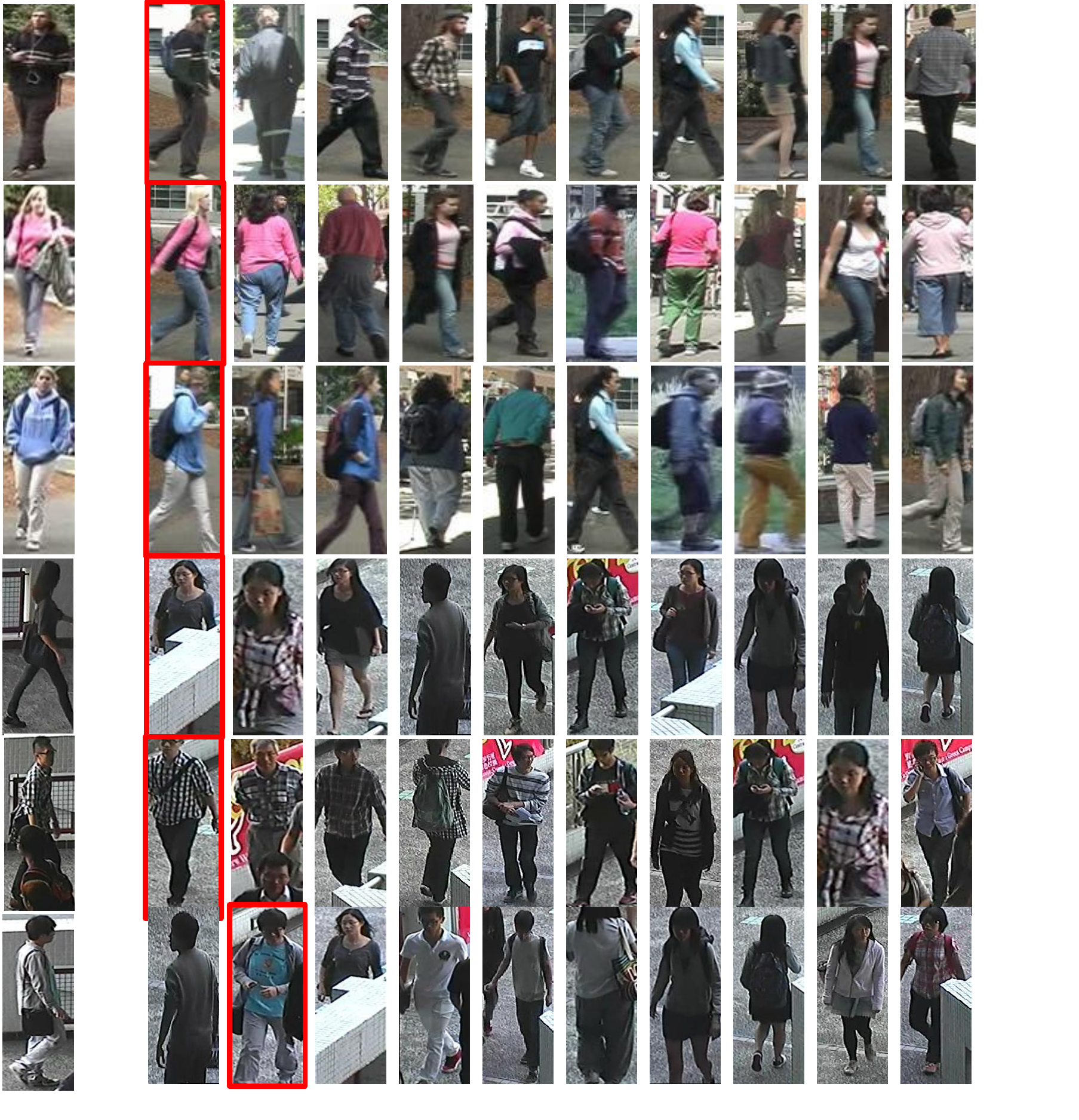}}
	\caption{Qualitative evaluation on VIPeR (first three rows) and CUHK-03 datasets. The single-shot setting with only one correct gallery image is more challenging for person re-ID. Red boxes indicate that the identity is same as that of query.}
	\label{fig:viperqual}
\end{figure}
\subsection{Comparison of backbone CNN architecture}
The base network for deep representation learning contains convolutional layers followed by Inception modules. Table~\ref{tab:architecturecomp} shows the effect of using ResNet-50 architecture~\cite{he2016deep} for our task of multi-body representation. ResNet-50 is deeper and takes a larger input image size of 224$\times$224 for learning the feature representations. However, this does not equate to a better representation for the unseen test images as the mAP measures are consistently better for both the datasets under consideration.

\subsection{Qualitative comparison}
Fig.~\ref{fig:marketdukequal} and \ref{fig:viperqual} show the retrieval results obtained using queries from all the datasets. For Market-1501 and DukeMTMC-reID, the gallery contains multiple images of the query identity captured from a camera different to that of the query. As can be seen from Fig.~\ref{fig:marketdukequal}, our method is able to capture most person belonging to the probe, despite a huge variation in appearance and pose. The single-shot setting followed by VIPeR and CUHK-03, where only a single instance of the query identity is present in the gallery is more difficult for person re-ID. Although the person is partially occluded (row 5 in Fig.~\ref{fig:viperqual}) or the appearance changes drastically (last row in Fig.~\ref{fig:viperqual}), our method is consistently able to retrieve the correct person.     

\section{Conclusion} 
In this paper, we have shown that deep classification models that are trained on the global image tend to learn poor representations for unseen test images. It is shown that a fusion of handcrafted features and deep feature representation learned using multiple body parts to complement the global body features achieves high performance on such zero-shot learning problems. Experimental evaluations on four benchmark datasets in person re-ID show that the proposed method performs among the state-of-the-art. 
%-------------------------------------------------------------------------

{\small
\bibliographystyle{ieee}
\bibliography{egbib}
}

\end{document}